\documentclass{article}

\PassOptionsToPackage{numbers, compress}{natbib}

 \usepackage[final]{neurips_2024}

\usepackage{multirow}

\usepackage[utf8]{inputenc} 
\usepackage[T1]{fontenc}    
\usepackage{hyperref}       
\usepackage{url}            
\usepackage{booktabs}       
\usepackage{amsfonts}       
\usepackage{amsmath}
\usepackage{nicefrac}       
\usepackage{microtype}      
\usepackage{xcolor}         
\usepackage{graphicx}
\usepackage{tikz}
\usepackage{caption}
\usepackage{subcaption}

\title{Hallo: Hierarchical Audio-Driven Visual Synthesis for Portrait Image Animation}

\author{
Mingwang Xu$^{1*}$ \quad 
Hui Li$^{1*}$ \quad 
Qingkun Su$^{1*}$ \quad 
Hanlin Shang$^1$ \quad 
Liwei Zhang$^1$ \quad
\textbf{Ce Liu}$^3$ \\
\textbf{Jingdong Wang}$^2$ \quad  
\textbf{Yao Yao}$^4$ \quad 
\textbf{Siyu Zhu}$^{1}$ \\
$^1$Fudan University \quad 
$^2$Baidu Inc \quad 
$^3$ETH Zurich \quad 
$^4$Nanjing University
}

\begin{document}

\maketitle

\begin{abstract}
The field of portrait image animation, driven by speech audio input, has experienced significant advancements in the generation of realistic and dynamic portraits. 
This research delves into the complexities of synchronizing facial movements and creating visually appealing, temporally consistent animations within the framework of diffusion-based methodologies. 
Moving away from traditional paradigms that rely on parametric models for intermediate facial representations, our innovative approach embraces the end-to-end diffusion paradigm and introduces a hierarchical audio-driven visual synthesis module to enhance the precision of alignment between audio inputs and visual outputs, encompassing lip, expression, and pose motion. 
Our proposed network architecture seamlessly integrates diffusion-based generative models, a UNet-based denoiser, temporal alignment techniques, and a reference network. 
The proposed hierarchical audio-driven visual synthesis offers adaptive control over expression and pose diversity, enabling more effective personalization tailored to different identities.
Through a comprehensive evaluation that incorporates both qualitative and quantitative analyses, our approach demonstrates obvious enhancements in image and video quality, lip synchronization precision, and motion diversity. 
Further visualization and access to the source code can be found at: \url{https://fudan-generative-vision.github.io/hallo}.
\end{abstract}

\let\thefootnote\relax\footnotetext{$^*$ indicates equal contribution.}

\begin{figure}[!t]
    \centering
    \includegraphics[width=1.0\linewidth]{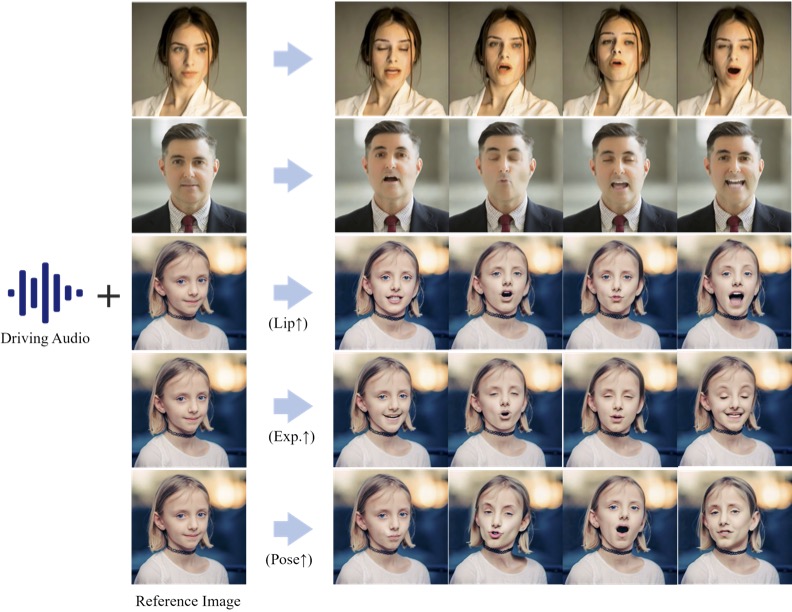}
    \caption{
The proposed methodology aims to generate portrait image animations that are temporally consistent and visually high-fidelity.
This is achieved by utilizing a reference image, an audio sequence, and optionally, the visual synthesis weight in conjunction with a diffusion model based on the hierarchical audio-driven visual synthesis approach. 
The results of this method showcase improved fidelity and visual quality in comparison to previous approaches that rely on intermediate facial representations. 
Furthermore, the proposed methodology enhances the accuracy of lip synchronization measurement and enhances the control over motion diversity.
}
\label{fig:teaser}
\end{figure}

\section{Introduction}

Portrait image animation, also known as talking head image animation, aims to generate a speaking portrait from a single static image and a corresponding speech audio.
This technology holds substantial value across various domains such as video gaming and virtual reality, film and television production, social media and digital marketing, online education and training, as well as human-computer interaction and virtual assistants.
Significant advancements in this field are exemplified by works like Stable Diffusion~\cite{rombach2021highresolution} and DiT~\cite{Peebles2022DiT}, which involve the gradual incorporation of noise into the training data in the latent space, followed by a reverse process that progressively reconstructs the signal from this noise. 
By leveraging parametric or implicit representations of lip movements~\cite{prajwal2020lip,zhang2023sadtalker}, facial expressions~\cite{he2023gaia,ma2023dreamtalk,shen2023difftalk}, and head poses~\cite{ma2023dreamtalk,stypulkowski2024diffused,sun2023vividtalk}, combined with diffusion techniques in the latent space~\cite{corona2024vlogger,liu2024anitalker,tian2024emo,wei2024aniportrait,xu2024vasa}, it is possible to generate high-quality, lifelike dynamic portraits in an end-to-end manner.

This study focuses on two primary challenges in portrait image animation:
(1) The synchronization and coordination of lip movements, facial expressions, and head poses driven by the audio input.
(2) The creation of high-fidelity animations that are visually appealing and maintain temporal consistency.
Addressing the first challenge, parametric model-based approaches~\cite{corona2024vlogger,ma2023dreamtalk,prajwal2020lip,shen2023difftalk,sun2023vividtalk,wei2024aniportrait,zhang2023sadtalker} initially rely on input audio to derive sequences of facial landmarks or parametric motion coefficients, such as 3DMM~\cite{blanz2003face}, FLAME~\cite{li2017learning} and HeadNeRF~\cite{hong2022headnerf}, as intermediate representations, which are subsequently used for visual generation. 
However, the effectiveness of these methods is constrained by the accuracy and expressiveness of the intermediate representations.
In contrast, another approach involves decoupled representation learning in latent space~\cite{he2023gaia,liu2024anitalker,xu2024vasa}, which segregates facial features into identity and non-identity components within a latent space. 
These components are independently learned and integrated across frames. 
The main obstacle in this approach is effectively disentangling the various portrait factors while encompassing static and dynamic facial attributes comprehensively.
Moreover, the introduction of intermediary representations or decoupled representations in the latent space of the aforementioned methods undermine the production of temporally consistent visual outputs with high realism and diversity.

To generate visually appealing and temporally consistent high-quality animations, the utilization of end-to-end diffusion models~\cite{tian2024emo} leverages recent advancements in diffusion models for synthesizing portrait videos directly from images and audio clips. 
This approach eliminates the need for intermediate representations or complex preprocessing steps, enabling the production of high-resolution, detailed, and diverse visual outputs. 
However, achieving precise alignment between audio and the generated facial visual synthesis, including lip movements, expressions, and poses, still remains a challenge.
In this study, we adhere to the framework of end-to-end diffusion models with the objective of addressing the alignment issue while incorporating the ability to control the diversity of expressions and poses. 
To tackle this challenge, we introduce a hierarchical audio-driven visual synthesis module that employs cross-attention mechanisms to establish correspondences between audio and visual features associated with lips, expressions, and poses.
Subsequently, these cross-attentions are fused using adaptive weighting.
Building upon this hierarchical audio-driven visual synthesis module, we propose a network architecture that integrates a diffusion-based generative model with a UNet-based denoiser~\cite{rombach2022highresolution}, temporal alignment~\cite{guo2024animatediff} for sequence coherence, and a ReferenceNet to guide visual generation.
As shown in Figure~\ref{fig:teaser}, this integrated approach enhances lip and expression movements to synchronize effectively with the audio signal and offers adaptive control over the granularity of expressions and poses, thereby ensuring consistency and realism across various visual identities.


In the experimental section, our proposed approach is comprehensively evaluated from both qualitative and quantitative perspectives, encompassing assessments of image and video quality, lip synchronization, and motion diversity.
Specifically, our method demonstrates a significant enhancement in image and video quality compared to previous methodologies, as quantified by the FID and FVD metrics.
Moreover, our approach exhibits a promising advancement in lip synchronization when contrasted with previous diffusion-based techniques.
Furthermore, our method offers the flexibility to adjust the diversity of expressions and poses according to specific requirements.
It is noteworthy that we are dedicated to the dissemination of our source code and sample data to the open-source community, with the aim of facilitating further research in related fields.

\section{Related Work}

\paragraph{Diffusion-Based Video Generation.}
Recent advancements in diffusion-based video generation have shown promising results~\cite{blattmann2023align, dai2023animateanything, harvey2022flexible, ho2022video, khachatryan2023text2videozero, luo2023videofusion, yang2022diffusion,zhu2024champ}, leveraging the foundational principles of text-to-image diffusion models. 
Notable efforts include Video Diffusion Models (VDM)~\cite{ho2022video}, which utilize a space-time factorized U-Net for simultaneous image and video data training, and ImagenVideo~\cite{ho2022imagen}, which enhances VDM with cascaded diffusion models for high-definition outputs. 
Make-A-Video~\cite{singer2022makeavideo} and MagicVideo~\cite{zhou2023magicvideo} extend these concepts to facilitate seamless text-to-video transformations. 
For video editing, techniques like VideoP2P~\cite{liu2023videop2p}  and vid2vid-zero~\cite{wang2024zeroshot} manipulate cross-attention maps to refine outputs, while Dreamix~\cite{molad2023dreamix} employs image-video mixed fine-tuning. 
Additionally, Gen-1~\cite{esser2023structure} integrates structural guidance through depth maps and cross-attention, whereas MCDiff~\cite{chen2023motionconditioned} and LaMD~\cite{hu2023lamd} focus on motion-guided video generation, enhancing the realism of human actions. 
VideoComposer~\cite{wang2023videocomposer}, AnimateDiff~\cite{chen2023videocrafter1}, and VideoCrafter~\cite{chen2023videocrafter1} further explore the synthesis of image-to-video generation by conditioning the diffusion process on images and blending image latents. 
Given the remarkable efficacy of diffusion models in achieving high-fidelity and temporally consistent visual generation outcomes, an end-to-end diffusion model is adopted in this study.

\paragraph{Facial Representation Learning.}
Learning facial representations in the latent space, particularly disentangled from identity-related appearances and non-identity-related motions such as lip movements, expressions, and poses, represents a significant challenge in the field of computer vision. 
Representation learning related to this challenge generally falls into two categories: explicit and implicit methods.
Explicit methods often employ facial landmarks, which are key points~\cite{siarohin2019first,zakharov2020fast} on the face used for localization and representation of critical regions such as the mouth, eyes, eyebrows, nose, and jawline. 
Additionally, work involving 3D parametric models~\cite{gao2023high,ren2021pirenderer,zhang2023metaportrait}, like the 3D morphable Model (3DMM), captures the variability in human faces through a statistical shape model coupled with a texture model. 
However, these explicit methods are limited by their expressive capabilities and the precision of reconstructions.
Conversely, implicit methods aim to learn disentangled representations in 2D~\cite{burkov2020neural,liang2022expressive,pang2023dpe,wang2023progressive,yin2022styleheat,zhou2021pose} or 3D~\cite{drobyshev2022megaportraits,wang2021one} latent spaces, focusing on aspects such as appearance identity, facial dynamics, and head pose. 
While these approaches have yielded promising results in expressive facial representations, they face similar challenges to explicit methods, notably the accurate and effective disentanglement of various facial factors remains a considerable challenge.

\paragraph{Portrait Image Animation.}
In recent years, the field of portrait image animation has made substantial progress, particularly in generating realistic and expressive talking head animations from static images paired with audio inputs. 
The advancement began with LipSyncExpert~\cite{prajwal2020lip}, which improved lip synchronization to speech segments, achieving high accuracy for static images and videos of specific identities. 
This was followed by developments such as SadTalker~\cite{zhang2023sadtalker} and DiffTalk~\cite{shen2023difftalk}, which integrated 3D information and control mechanisms to enhance the naturalism of head movements and expressions.
A major shift occurred with the introduction of diffusion models, notably Diffused Heads~\cite{stypulkowski2024diffused} and GAIA~\cite{he2023gaia}, which facilitated the creation of high-fidelity talking head videos featuring natural head motion and facial expressions without reliance on additional reference videos. 
These models demonstrated robust capabilities for personalized and generalized synthesis across various identities.
Continuing this evolution, DreamTalk~\cite{ma2023dreamtalk} and VividTalk~\cite{sun2023vividtalk} utilized diffusion models to produce expressive, high-quality audio-driven facial animations, emphasizing improved lip synchronization and diverse speaking styles. 
Further advancements were made by Vlogger~\cite{corona2024vlogger} and AniPortrait~\cite{wei2024aniportrait}, which introduced spatial and temporal controls to accommodate variable video lengths and customizable character representations, respectively.
Recent innovations such as VASA-1~\cite{xu2024vasa} and EMO~\cite{tian2024emo} have developed frameworks for creating emotionally expressive and realistic talking faces from static images and audio, capturing a broad spectrum of facial nuances and head movements. 
Lastly, AniTalker~\cite{liu2024anitalker} introduced a groundbreaking framework focused on generating detailed and realistic facial movements through a universal motion representation, minimizing the need for labeled data and highlighting the potential for dynamic avatar creation in practical applications. 
In this study, we adopt the latent diffusion formulation as espoused in EMO~\cite{tian2024emo} and introduce a hierarchical cross-attention mechanism to augment the correlation between audio inputs and non-identity-related motions such as lip movements, expressions, and poses.
Such a formulation not only affords adaptive control over expression and pose diversity but also enhances the overall coherence and naturalness of the generated animations.

\begin{figure}[!t]
    \centering
    \includegraphics[width=1\linewidth]{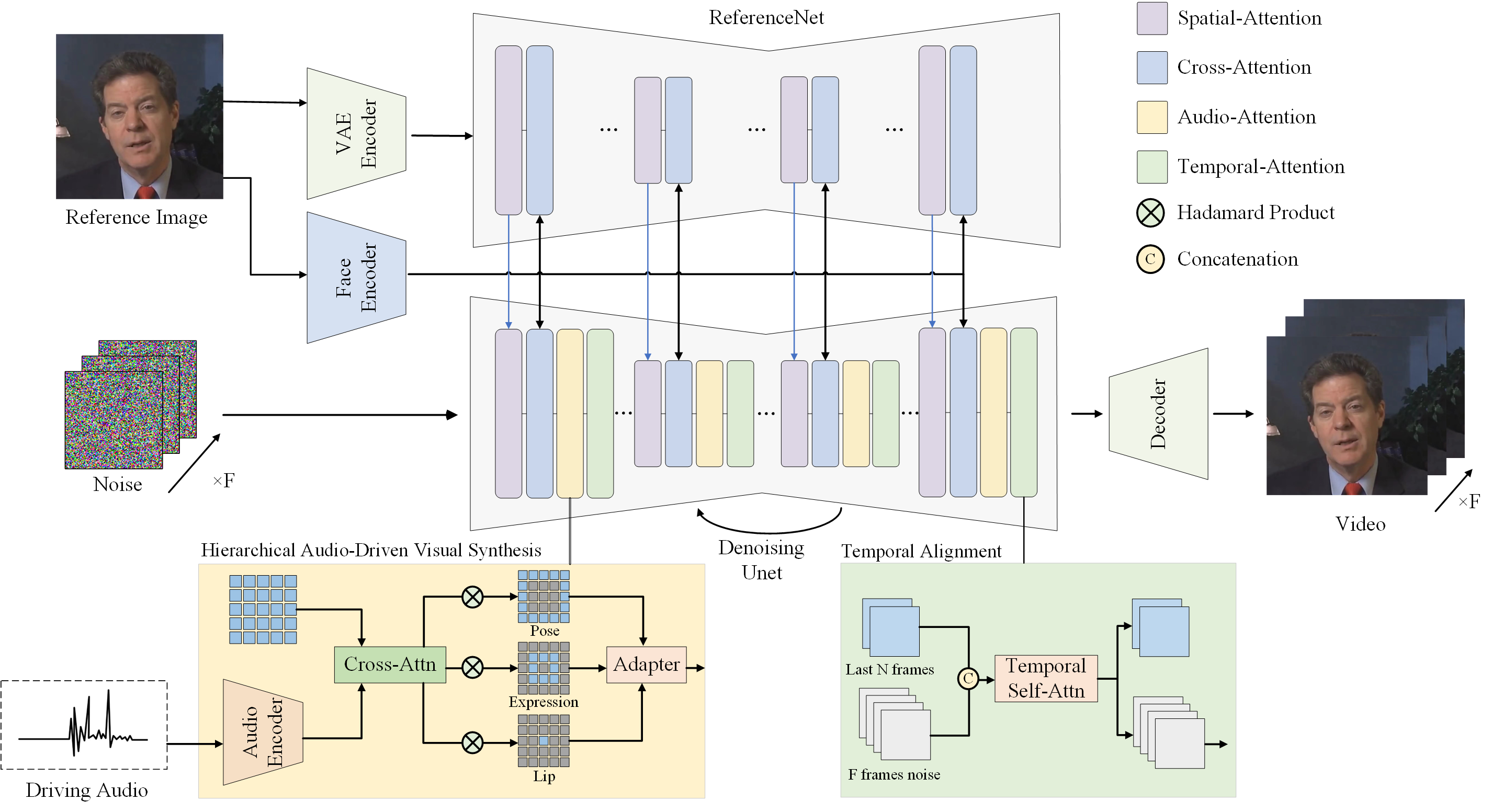}
    \caption{
The overview of the proposed pipeline.
Specifically, we integrates a reference image containing a portrait with corresponding audio input to drive portrait animation. 
Optional visual synthesis weights can be used to balance lip, expression, and pose weights. 
ReferenceNet encodes global visual texture information for consistent and controllable character animation. 
Face and audio encoders generate high-fidelity portrait identity features and encode audio as motion information respectively. 
The module of hierarchical audio-driven visual synthesis establishes relationships between audio and visual components (lips, expression, pose), with a UNet denoiser used in the diffusion process.
}
 \label{fig:pipeline}
\end{figure}

\section{Methodology}

\subsection{Preliminaries}
\paragraph{Latent Diffusion Models.}
Latent diffusion models represent a class of diffusion models that operate within the encoded latent space produced by an autoencoder, formalized as $\mathcal{D}(\mathcal{E}(\cdot))$. 
A prominent instance of an latent diffusion model is Stable Diffusion, which integrates a VQ-VAE autoencoder~\cite{van2017neural} and a time-conditioned U-Net for noise estimation. 
Moreover, Stable Diffusion~\cite{rombach2022highresolution} employs a CLIP ViT-L/14 text encoder~\cite{radford2021learning} to transform the input text prompt into a corresponding text embedding, which serves as a condition during the diffusion process.
During the training phase, given an image $I\in\mathbb{R}^{H_I\times W_I \times 3}$ and its associated text condition $c_\text{embed}\in\mathbb{R}^{D_c}$, the latent representation:
\begin{equation} 
z_0 = \mathcal{E}(I)\in\mathbb{R}^{H_z\times W_z \times D_z}
\end{equation} 
undergoes a diffusion process across $T$ timesteps.
This process is modeled as a deterministic Gaussian process, culminating in $z_T \sim \mathcal{N}(0, I)$.
The training objective for Stable Diffusion is encapsulated by the following loss function: 
\begin{equation} 
L = \mathbb{E}_{\mathcal{E}(I), c_{\text{embed}},\epsilon\sim\mathcal{N}(0,1),t}\left[\lVert \epsilon-\epsilon_{\theta}(z_t, t, c_{\text{embed}}) \rVert_{2}^{2} \right], 
\end{equation} 
with $t$ uniformly sampled from $\{1,...,T\}$. Here, $\epsilon_{\theta}$ denotes the trainable components of the model, which include a denoising U-Net equipped with Residual Blocks and layers that facilitate both self-attention and cross-attention. 
These components are tasked with processing the noisy latent variables $z_t$ and the conditional embeddings $c_{\text{embed}}$.

Upon the completion of training, the original latent $z_t$ is reconstructed using deterministic sampling methods such as denoising diffusion implicit models (DDIM)~\cite{radford2021learning}. 
The latent $z_t$ is then decoded by the decoder $\mathcal{D}$ to generate the final image output. 
This methodology not only preserves the fidelity of the generated images but also ensures that they are contextually aligned with the input auxiliary conditions, demonstrating the efficacy of integrating conditioned processes in generative models.

\paragraph{Cross Attention as Motion Guidance.}
Cross attention is a critical component of the latent diffusion model framework, significantly enhancing the generative process by directing the flow and focus of information. 
This mechanism allows the model to concentrate on specific motion guidance and latent image representations, thus improving the semantic coherence between the generated videos and the motion condition. 
In general video generation tasks, motion conditions may include textual prompts or dense motion flows. In more specialized domains such as human image animation, motion conditions may involve skeletons, semantic maps, and 3D parametric flows.
For portrait image animation tasks, audio is typically used as the motion condition, with cross attention employed to integrate the relevant conditions effectively.
Mathematically, cross attention is implemented using attention layers that process both the embedding of the motion condition, denoted as $c_{\text{embed}}$, and the noisy latent variables $z_t$.
These layers calculate attention scores that determine the model's focus level on various input aspects. 
The scores are then utilized to weight the contributions of each component, creating a weighted sum that serves as the input for subsequent layers. 
The cross attention mechanism can be expressed as:
\begin{equation}
\text{CrossAttn}(z_t, c_{\text{embed}}) = \text{Attention}(Q, K, V).
\end{equation}
Here, $\text{CrossAttn}(z_t, c_{\text{embed}})$ computes the attention scores using:
\begin{align}
Q &= W_Q \cdot z_t, \;\; W_Q\in\mathbb{R}^{D_e\times D_z}\\
K &= W_K \cdot c_{\text{embed}}, \;\; W_K \in\mathbb{R}^{D_e \times D_c}\\
V &= W_V \cdot c_{\text{embed}}, \;\; W_V\in\mathbb{R}^{D_e \times D_c}
\end{align}
where $W_Q$, $W_K$, and $W_V$ are learnable projection matrices.

The output is a weighted representation that captures the most pertinent aspects of the input condition and latent image at the current stage of the diffusion process. 
By incorporating cross attention, the latent diffusion model dynamically adjusts its focus based on the evolving state of the latent variables and the specified motion condition.

\begin{figure}[!t]
    \centering
        \includegraphics[width=\textwidth]{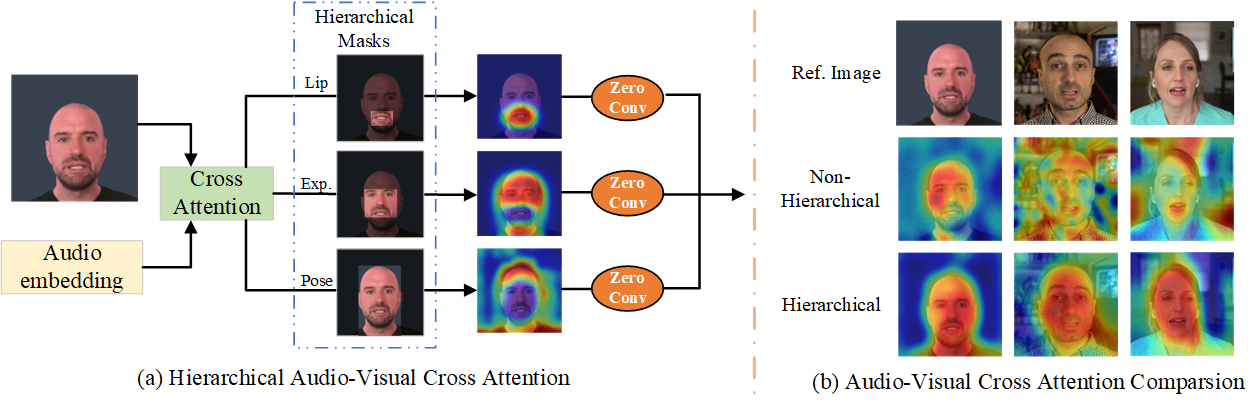}
    \caption{Visualization of the hierarchical audio-driven visual synthesis and a comparative analysis of the audio-visual cross attention between the original full method and our proposed hierarchical audio-visual cross attention.}
\label{fig:hadvs}
\end{figure}

\subsection{Hierarchical Audio-Driven Visual Synthesis.}
As demonstrated in Figure~\ref{fig:pipeline}, our task is to generate a video with $S$ frames given a face image and a clip of audio. For each frame $s$, we denote the corresponding latent representation at the diffusion time step $t$ as $z^{(s)}_t$. 
\paragraph{Face Embedding.}
The objective of face embedding is to produce high-fidelity portrait identity features. 
In contrast to previous approaches utilizing CLIP~\cite{radford2021learning} for generic visual feature encoding by jointly training on a large dataset of images and textual descriptions, our method employs a pre-trained face encoder to extract identity features $c_{\text{exp}}\in\mathbb{R}^{D_f}$. 
These features are fed into a diffusion network's cross-attention module, i.e., CrossAttn($z_t^{(s)}$, $c_{\text{exp}}$), to interact with latents, generating portrait animations faithful to the input character features. 
This approach not only ensures generalization in facial feature extraction but also accurately preserves and reproduces individual identity traits such as facial expressions, age, and gender.

\paragraph{Audio Embedding.}
To enhance the encoding of audio for motion-driven information in animation, we employed wav2vec~\cite{schneider2019wav2vec} as the audio feature encoder. 
Specifically, we concatenated the audio embeddings from the final 12 layers of the wav2vec network to capture richer semantic information across different audio layers. 
Given the contextual influence on sequential audio data, we extracted the corresponding 5-second audio segment for the $S$ frames. 
Utilizing three simple linear layers, we transform the pre-trained model's audio embeddings to $\{c_{\text{audio}}^{(s)}\}_{s=1}^S$, where $c_{\text{audio}}^{(s)}\in\mathbb{R}^{D_a}$ is the audio embedding for the $s$ frame.   

\paragraph{Hierarchical Audio-Visual Cross Attention.}
As shown in Figure~\ref{fig:hadvs}, we introduce a hierarchical audio-visual cross attention mechanism to facilitate the model's learning of relationships between audio and visual components such as lips, expression and pose.

First, we pre-process the face image to obtain masks $M_{\text{lip}}, M_{\text{exp}}, M_{\text{pose}} \in\{0,1\}^{H_z\times W_z}$ representing the lip, expression and pose areas respectively. 
Specifically, we use the off-the-shelf toolbox MediaPipe to predict landmarks from the face image $I$. The MediaPipe predicts the coordinates of multiple landmarks for lip and expression. 
The bounding box masks $Y_\text{lip}$, $Y_\text{exp}\in\{0,1\}^{H_z\times W_z}$ are obtained by
\begin{equation}
    Y_{\text{r}}(i,j)= 
    \begin{cases}
    1 &\text{if $(i,j)$ is inside $\text{BoundingBox(r)}$}\\
    0 &\text{otherwise}
    \end{cases},
\end{equation}
where $r\in\{\text{lip},\text{exp}\}$, and $\text{BoundingBox(r)}$ denotes the bounding box of all the landmarks for $\text{r}$. 
We compute
\begin{align}
M_{\text{lip}} &= Y_\text{lip}, \\
M_{\text{exp}} &= (1-M_{\text{lip}})\odot Y_\text{exp}, \\
M_{\text{pose}} &= 1-M_{\text{exp}},
\end{align}
where $\odot$ is the Hadamard product.

Next, we apply the cross-attention mechanism between the latent representations and the audio embeddings:
\begin{equation}
o_t^{(s)} = \text{CrossAttn}(z_t^{(s)}, c_{\text{audio}}^{(s)}).
\end{equation}
We apply masks and obtain different scaled latent representations by
\begin{align}
 b_t^{(s)} &= o_t^{(s)}\odot M_{\text{pose}}, \\
 f_t^{(s)} &= o_t^{(s)}\odot M_{\text{exp}}, \\
 l_t^{(s)} &= o_t^{(s)}\odot M_{\text{lip}}.
 \end{align}

In the end, to effectively merge these outputs, we add a adaptive module to handle hierarchical audio guidance. Specifically, the output is obtained by the convolution layer:
\begin{equation}
\sum_u \text{Conv}_u(u), \; u\in\{a_t^{(s)}, f_t^{(s)}, l_t^{(s)}\}.
\end{equation}



\subsection{Network Architecture}

Finally, we present the network architecture of the diffusion-based generative model with the hierarchical audio-visual cross-attention proposed in this paper.
We employed a standard visual generative model for video generation based on Stable Diffusion, utilizing ReferenceNet to guide the generation process using existing reference images.
Temporal alignment enhances coherence and consistency in the generated video sequence over time, in addition to the hierarchical audio-driven visual synthesis module that establishes refined mappings between audio and lip, expression, and pose, as previously described.

\paragraph{Diffusion Backbone.}
Stable Diffusion 1.5 is built upon a latent diffusion model that consists of three primary components: Vector Quantised Variational AutoEncoder (VQ-VAE), Unet-based denoising model, and a conditioning module. 
In the context of text-to-image applications, latent image inputs are generated from random initialization and processed by the diffusion model in collaboration with the conditioning module to generate new latent image outputs.
In this study, text features are excluded from the conditioning in Stable Diffusion, as audio signals are employed as the primary driver for motion.

\paragraph{ReferenceNet.}
ReferenceNet is designed to guide the generation process by referencing existing images to enhance the quality of generated videos, including visual texture information of portraits and backgrounds.
It is a Unet-based Stable Diffusion network with the same number of layers as the denoising model network.
The feature maps generated by these structures at specific layers are likely to exhibit similarities, aiding in the integration of extracted features by ReferenceNet into the diffusion backbone.
These structures generate feature maps at specific layers with the same spatial resolution and potentially similar semantic features. 
This integration enhances the quality of generated videos, including visual texture information of portraits and backgrounds. 
During model training, the first frame of the video clip serves as the reference image.

\paragraph{Temporal Alignment.}
In video generation tasks based on diffusion models, temporal alignment plays a crucial role in ensuring the coherence and consistency of the generated video sequences over time. 
More precisely, a subset of frames (we use 2 in our implementation) from the preceding inference step are designated as motion frames, which are then concatenated with the latent noise and manipulated along the temporal axis. 
This temporal manipulation is facilitated through multiple self-attention blocks, each adept at processing the temporal sequence elements of the video frames.

\begin{figure}[!t]
    \begin{minipage}{0.53\textwidth}
        \centering
        \includegraphics[width=\textwidth]{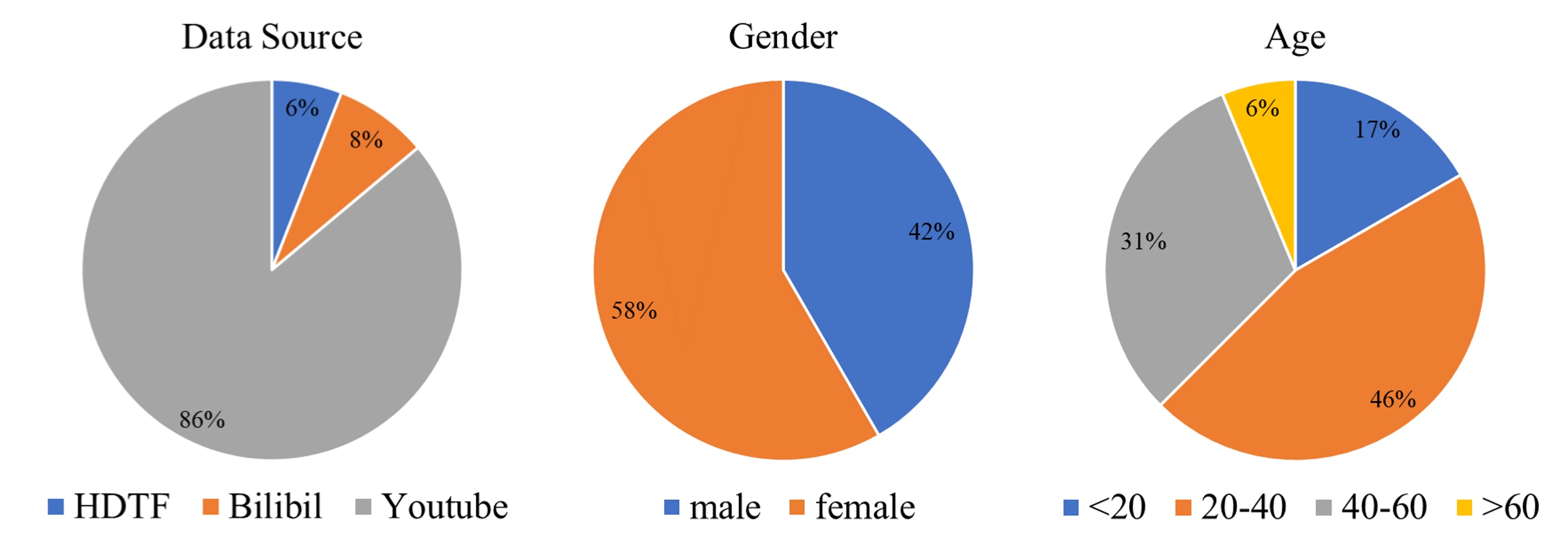}
    \end{minipage}  
    \;
    \begin{minipage}{0.40\textwidth}
        \centering
        \resizebox{1.15\columnwidth}{!}{
        \begin{tabular}{ccccc}
            \hline
            \multicolumn{1}{c}{\multirow{2}{*}{Datasets}} & \multicolumn{2}{c}{Raw} & \multicolumn{2}{c}{Filtered} \\ \cline{2-5} 
            \multicolumn{1}{c}{}                          & \#IDs     & \#Hours     & \#IDs        & \#Hours       \\ \hline
            HDTF                                          & 190       & 8.42        & 188          & 6.47          \\
            Bilibili                                      & 402       & 18.41       & 311          & 8.66          \\
            Youtube                                       & 1617      & 137.49      & 1324         & 93.73         \\ \hline
            Total                                         & 2209      & 164.32      & 1823         & 108.86        \\ \hline
        \end{tabular}}
    \end{minipage}
\caption{Statistics of the dataset for training and inference.}
\label{fig:statistics}
\end{figure}

\subsection{Training and Inference}

\paragraph{Training.}
The training process consists of two distinct stages:
(1) In the first training phase, individual video frames are generated by utilizing reference image and target video frame pairs. 
The parameters of VAE encoder and decoder, along with the facial image encoder, are fixed, while allowing the weights of the spatial cross-attention modules of ReferenceNet and denoising UNet to be optimized to improve single-frame generation capability. 
Video clips containing 14 frames are extracted for input data, with a random frame from the facial video clip chosen as the reference frame and another frame from the same video as the target image.
(2) In the second training phase, video sequences are trained using reference images, input audio, and target video data. 
The spatial modules of ReferenceNet and denoising UNet remain static, focusing on enhancing video sequence generation capability. 
This phase predominantly focuses on training hierarchical audio-visual cross-attention to establish the relationship between audio as motion guidance and the visual information of lip, expression, and pose.
Additionally, motion modules are introduced to improve model temporal coherence and smoothness, initialized with pre-existing weights from AnimateDiff~\cite{guo2024animatediff}. 
One frame from a video clip is randomly selected as the reference image during this phase.

\paragraph{Inference.}
During the inference stage, the network takes a single reference image and driving audio as input, producing a video sequence that animates the reference image based on the corresponding audio.
To produce visually consistent long videos, we utilize the last 2 frames of the previous video clip as the initial $k$ frames of the next clip, enabling incremental inference for video clip generation.

\begin{table}[!t]
\centering
\begin{tabular}{c|cc|cc|cc}
\hline
Method      & \multicolumn{1}{c}{FID$\downarrow$} & \multicolumn{1}{c|}{FVD$\downarrow$} & \multicolumn{1}{c}{Sync-C$\uparrow$} & \multicolumn{1}{l|}{Sync-D$\downarrow$} & \multicolumn{1}{c}{E-FID$\downarrow$} \\ \hline
SadTalker~\cite{zhang2022sadtalker}   & 22.340       & 203.860        & 7.885                      & 7.545                       & 9.776                    \\
Audio2Head~\cite{wang2021audio2head}  & 37.776                  & 239.860                  & \textbf{8.024}             & \textbf{7.145}              & 17.103              \\
DreamTalk~\cite{ma2023dreamtalk}   & 78.147                  & 790.660                  & 6.376                      & 8.364                       & 15.696                \\
AniPortrait~\cite{wei2024aniportrait} & 26.561                  & 234.666                  & 4.015                      & 10.548                      & 13.754                    \\
Ours    & \textbf{20.545}                  & \textbf{173.497}                  & 7.750                      & 7.659                       & \textbf{7.951}             \\ \hline
Real video    &  -                 &  -                &        8.700            &     6.597           & -            \\ \hline
\end{tabular}
\vspace{2mm}
\caption{
The quantitative comparisons with the existed portrait image animation approaches on the HDTF data-set.
Our proposed method excels in generating high-quality, temporally coherent talking head animations with superior lip synchronization performance.
}
\label{tab:quantitative_hdtf}
\end{table}

\begin{figure}[!t]
    \centering
    \includegraphics[width=1\linewidth]{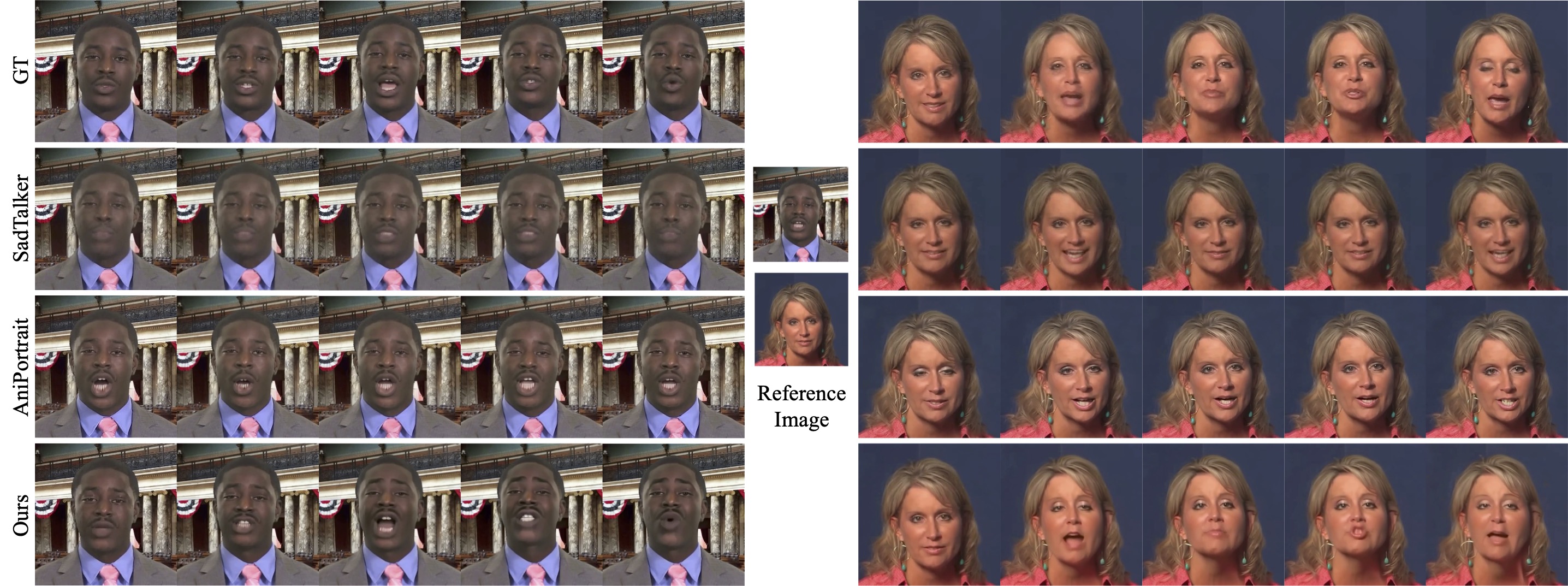}
    \caption{Qualitative comparison with existing approaches on HDTF data-set.}
    \label{fig:hdtf}
\end{figure}

\section{Experiment}
\subsection{Experimental Setups}
\paragraph{Implementation Details.}
Experiments encompassing both training and inference were carried out on a computational platform featuring 8 NVIDIA A100 GPUs. 
The initial and subsequent stages comprised 30,000 training steps each, with a batch size of 4 and video dimensions set at  $512\times512$.
Each training instance in the second stage produced 14 video frames, with the motion module's latents concatenated with the first 2 ground truth frames for video continuity. 
In both training stages, a learning rate of 1e-5 was employed, and the motion module was initialized with weights from Animatediff. 
To enhance video generation, the reference image, guidance audio, and motion frames were dropped with a 0.05 probability during training. 
In inference, continuity across sequences was ensured by concatenating noisy latents with feature maps of the last 2 motion frames from the previous step within the motion module.

\paragraph{Datasets.}
As shown in Figure~\ref{fig:statistics}, our dataset comprises HDTF (190 clips, 8.42 hours) and additional Internet-sourced data (2019 clips, 155.90 hours). 
As shown in Figure~\ref{fig:statistics}, these videos are consisted of individuals of diverse ages, ethnicities, and genders in half-body or close-up shots against varied indoor and outdoor backgrounds. 
To ensure high-quality training data, we underwent a data cleaning process that focused on retaining single-person speaking videos exhibiting strong lip and audio consistency, while excluding videos with scene changes, significant camera movements, excessive facial motion, and fully side-facing shots. 
Mediapipe was utilized to determine expression and lip activity ranges in training videos, forging the basis for constructing expression and lip masks employed in both training and inference. 
Following data cleaning, our refined training dataset includes HDTF (188 clips, 6.47 hours) and Internet-sourced data (1635 clips, 102.39 hours), with each training video clip comprising 15 frames at a resolution of $512\times512$.

\paragraph{Evaluation Metrics.}
The evaluation metrics utilized in the portrait image animation approach include Fréchet Inception Distance (FID), Fréchet Video Distance (FVD), Synchronization-C (Sync-C), Synchronization-D (Sync-D) and E-FID. 
Specifically, FID and FVD measure the similarity between generated images and real data, with lower values indicating better performance and thus more realistic outputs. 
Sync-C and Sync-D evaluate the lip synchronization of generated videos in terms of content and dynamics, with higher Sync-C and lower Sync-D scores denoting better alignment with the audio. 
E-FID assesses the quality of generated images based on the features extracted from the Inception network, offering a refined evaluation of fidelity. 

\begin{table}[!t]
\centering
\begin{tabular}{c|c|c|c|c|c}
\hline
Method                                & FID$\downarrow$ & FVD$\downarrow$  & Sync-C$\uparrow$ & Sync-D$\downarrow$ & E-FID$\downarrow$ \\ \hline
SadTalker~\cite{zhang2022sadtalker}   & 50.015          & 471.163          & 6.922            & \textbf{7.921}     & 95.194                 \\
Audio2Head~\cite{wang2021audio2head}  & 84.793          & 457.499          & 6.518            & 8.143              & 153.618                 \\
DreamTalk~\cite{ma2023dreamtalk}      & 109.011         & 988.539          & 5.709            & 8.743              & 153.450                 \\
AniPortrait~\cite{wei2024aniportrait} & 46.915          & 477.179          & 2.853            & 11.709             & 88.986                 \\
Ours                                  & \textbf{44.578} & \textbf{377.117} & \textbf{7.191}   & 7.984              & \textbf{78.495}              \\ \hline
Real video                            & -               & -                & 7.372            & 7.518              & -                 \\ \hline
\end{tabular}
\vspace{2mm}
\caption{The quantitative comparisons with the existing portrait image animation approaches on the CelebV data-set.}
\label{tab:quantitative_celebv}
\end{table}

\begin{figure}[!t]
    \centering
            \includegraphics[width=1\linewidth]{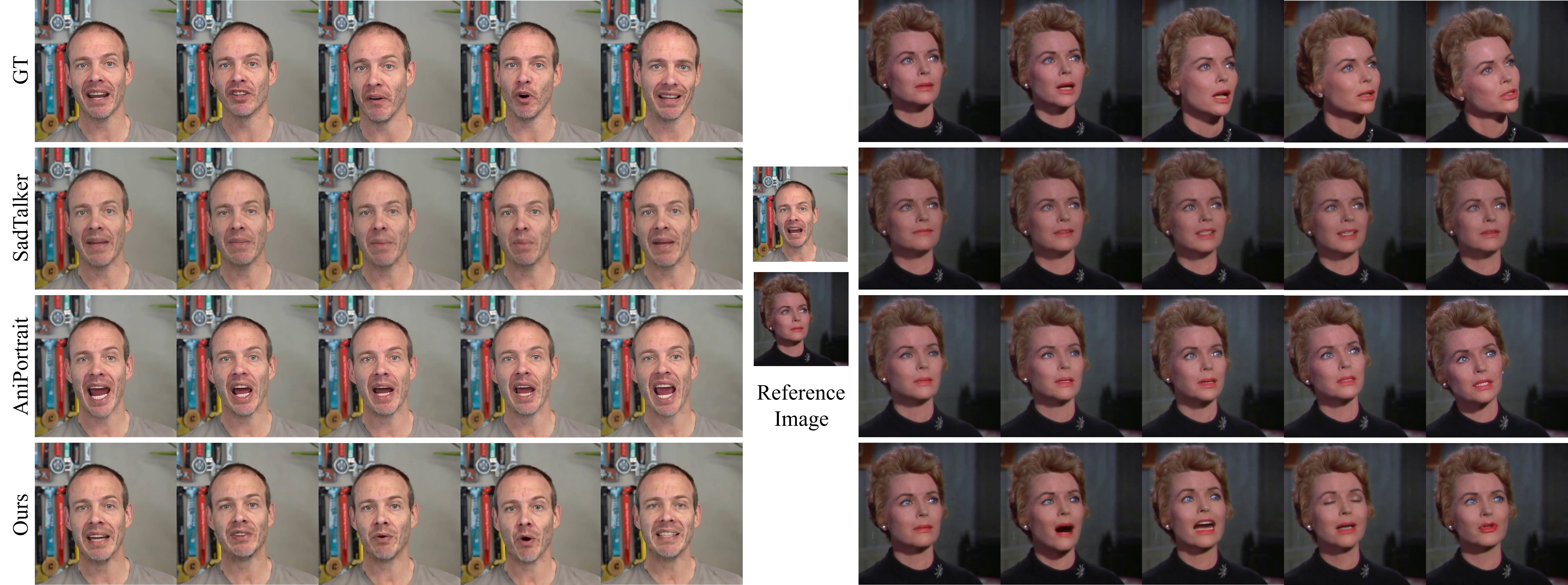}
    \caption{Qualitative comparison with existing approaches on CelebV data-set.}
    \label{fig:celebv}
\end{figure}

\paragraph{Baseline.}
In our quantitative experiments, we conducted a comparative analysis of our proposed method against publicly available implementations of SadTalker~\cite{zhang2022sadtalker}, DreamTalk~\cite{ma2023dreamtalk}, Audio2Head~\cite{wang2021audio2head}, and AniPortrait~\cite{wei2024aniportrait}. 
The evaluation was performed on the HDTF, CelebV and the proposed dataset, utilizing a training and testing split where 90\% of the identity data was allocated for training purposes.
The qualitative comparison encompassed an evaluation of our method against these selected approaches, taking into consideration reference images, audio inputs, and the resultant animated outputs provided by each respective method. 
This qualitative assessment aimed to provide insights into the performance and capabilities of our method in generating realistic and expressive talking head animations.

\begin{table}[!t]
\centering
\begin{tabular}{c|cc|cc|c}
\hline
Method                                                      & FID$\downarrow$ & FVD$\downarrow$  & Sync-C$\uparrow$ & Sync-D$\downarrow$ & E-FID$\downarrow$ \\ \hline
SadTalker~\cite{zhang2022sadtalker}   & 24.212 & 249.786 & 6.613            & 8.099     & 37.324                \\
Audio2Head~\cite{wang2021audio2head}  & 61.510          & 383.178          & 5.719            & 8.585              & 66.116                 \\
DreamTalk~\cite{ma2023dreamtalk}      & 128.423         & 964.088          & 5.925            & 8.596              & 58.180                 \\
AniPortrait~\cite{wei2024aniportrait} & 24.118          & 250.770          & 3.043            & 10.997             &  37.806                 \\
Ours                                                    & \textbf{23.266}         & \textbf{239.647}       & \textbf{6.924}   & \textbf{7.969}              & \textbf{34.731}                \\\hline
Real video                                                  & -               & -                &      7.011            &       7.606             & -                 \\ \hline
\end{tabular}
\vspace{2mm}
\caption{The quantitative comparisons with the existed portrait image animation approaches on the proposed ``wild'' data-set.}
\label{tab:quantitative_wild}
\end{table}

\begin{figure}[!t]
    \centering
    \includegraphics[width=0.9\linewidth]{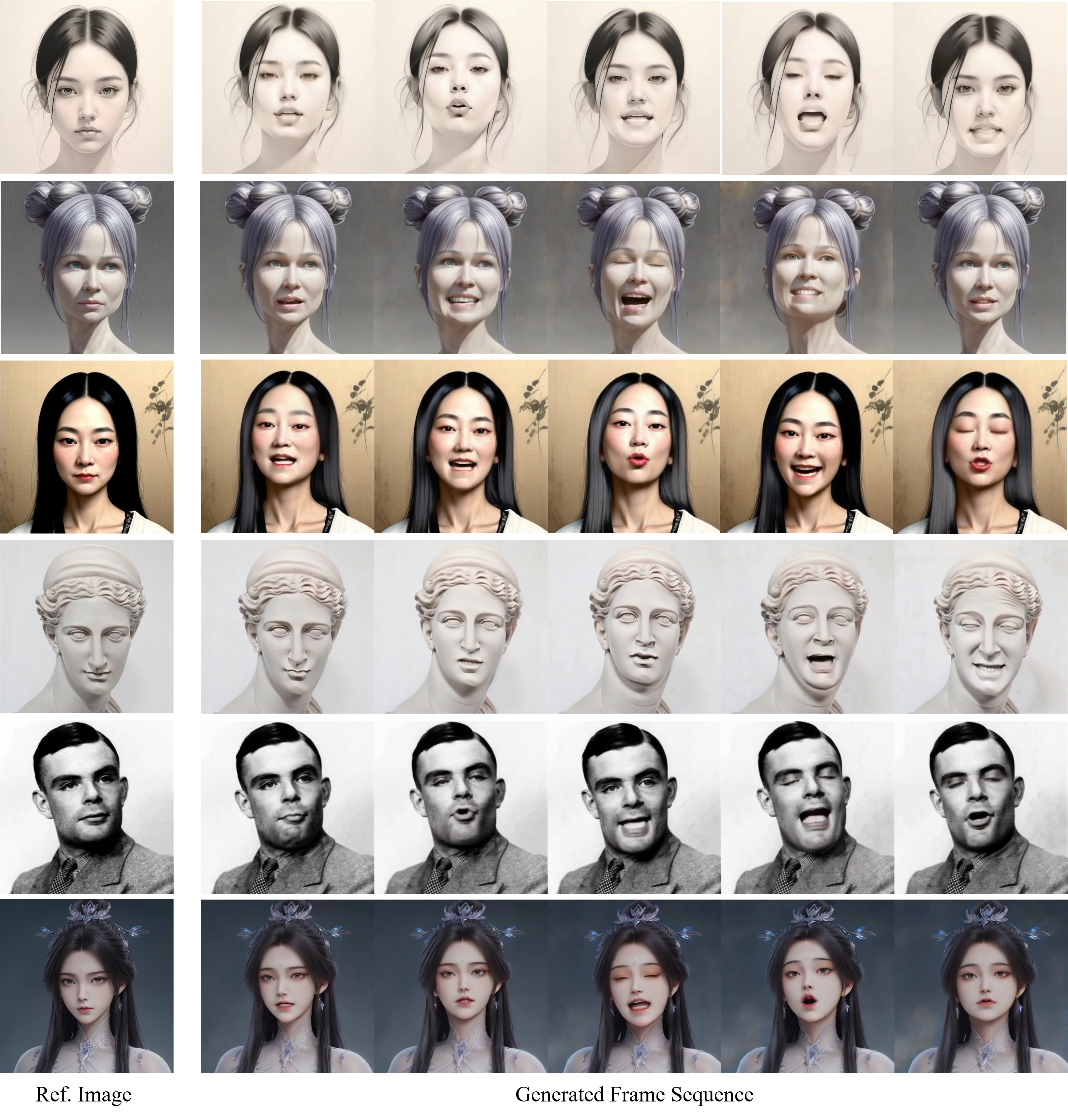}
    \caption{Video generation results of the proposed approach given different portrait styles.}
    \label{fig:qualitative_portrait}
\end{figure}

\subsection{Quantitative Results}
\paragraph{Comparison on HDTF Dataset.}
Table~\ref{tab:quantitative_hdtf} presents a comprehensive quantitative evaluation of various portrait image animation techniques on the HDTF dataset. 
Our proposed method demonstrates superior performance across several metrics, notably achieving the lowest FID (20.545), FVD (173.497), and E-FID (7.951). 
These results underscore the high quality and temporal coherence of the generated talking head animations. 
Additionally, our method exhibits commendable lip synchronization capabilities, as evidenced by the Sync-C (7.750) and Sync-D (7.659) metrics, which closely align with real video benchmarks. 
These achievements highlight the effectiveness of our approach in enhancing lip synchronization while maintaining high-fidelity visual generation and temporal consistency. 
To supplement these findings, Figure~\ref{fig:hdtf} visualizes the comparative performance of the different methods.

\begin{figure}[!t]
    \centering
    \includegraphics[width=1\linewidth]{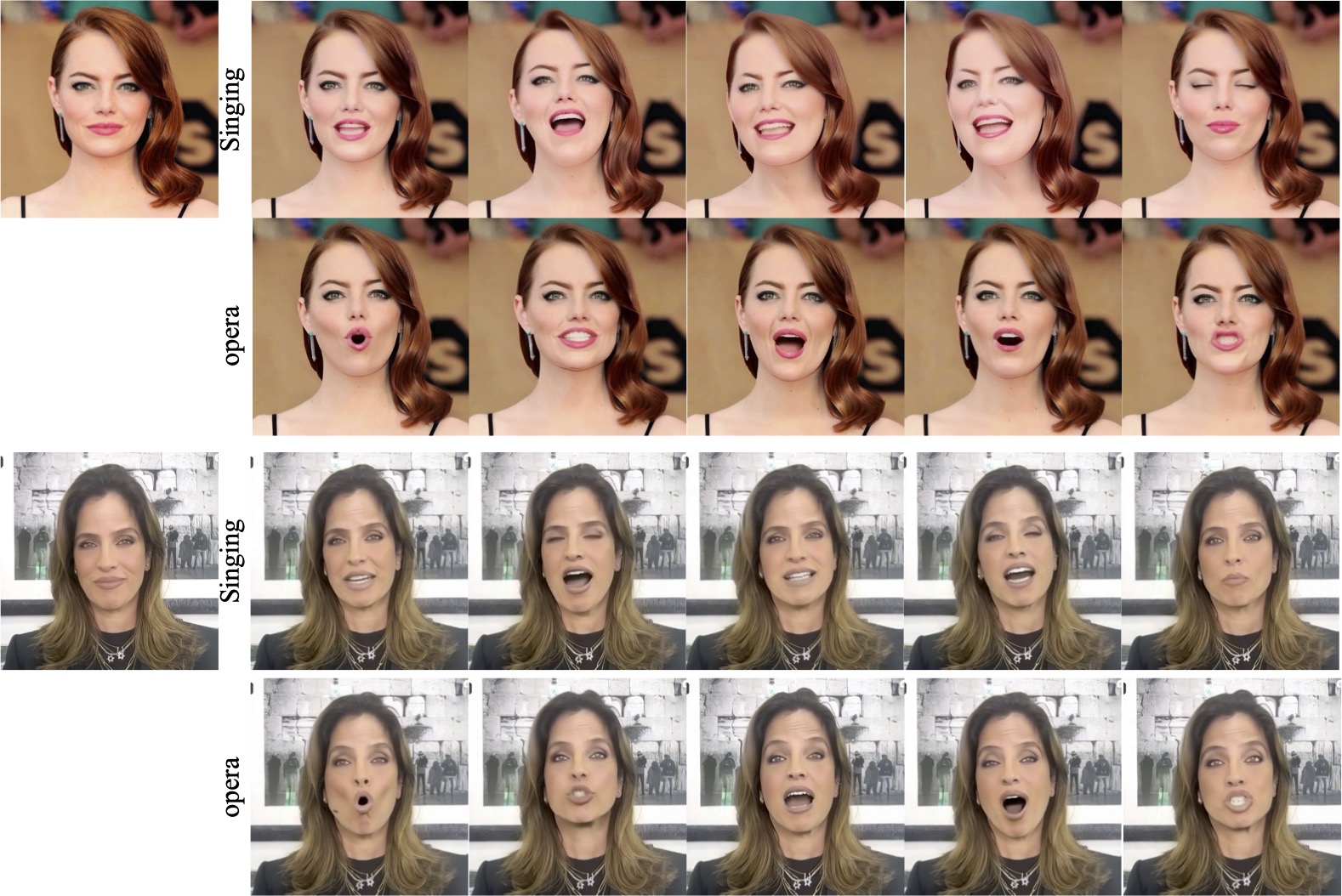}
    \caption{Video generation results of the proposed approach given different audio styles.}
    \label{fig:qualitative_audio}
\end{figure}

\begin{figure}[!t]
    \centering
    \includegraphics[width=1\linewidth]{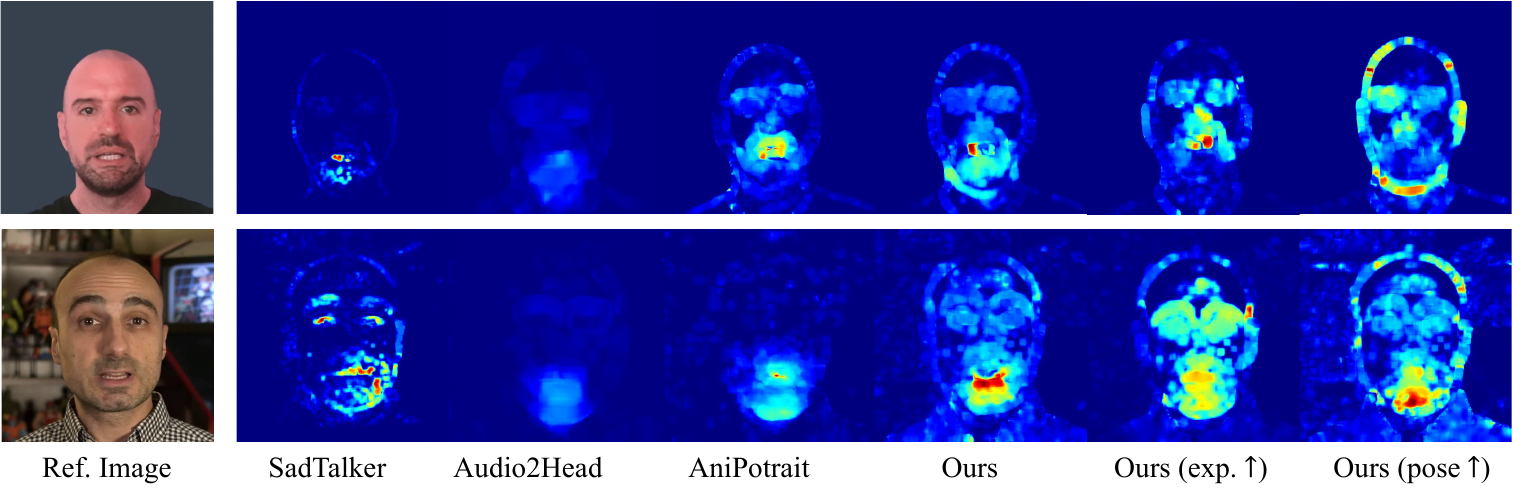}
    \caption{Qualitative comparison of head and expression motion diversity between existing methodologies and our proposed approach.}
    \label{fig:diversity}
\end{figure}

\paragraph{Comparison on CelebV Dataset.}
The quantitative evaluations presented in Table~\ref{tab:quantitative_celebv} provide a comparative analysis of various portrait image animation techniques using the CelebV dataset. 
Our proposed approach achieves superior performance, demonstrating the lowest FID (44.578) and FVD (377.117) scores, as well as the highest Sync-C score (7.191). 
Additionally, our method secures a competitive Sync-D value (7.984) and the lowest E-FID score (78.495), indicating high-quality animations with notable temporal coherence and precise lip synchronization. 
These results attest to the robustness and efficacy of our technique in generating realistic animations. To offer a more comprehensive analysis, a visual comparative examination is presented in Figure~\ref{fig:celebv}.

\paragraph{Comparison on the Proposed ``Wild'' Dataset.} 
The quantitative evaluations depicted in Table~\ref{tab:quantitative_wild} offer a comparative analysis of various portrait image animation techniques on the proposed ``wild'' dataset (as shown in statistics of Figure~\ref{fig:statistics} except for HDTF). 
Our methodology demonstrates outstanding performance, achieving the lowest FID (23.266) and FVD (239.647) scores, as well as the highest Sync-C score (6.924). 
Additionally, our approach attains the lowest E-FID score (34.731) and a competitive Sync-D value (7.969), closely approximating benchmarks set by real video data. 
These results highlight the robustness and efficacy of our technique in generating high-quality animations characterized by temporal coherence and precise lip synchronization under diverse and challenging conditions.

\begin{figure}[!t]
    \centering
    \includegraphics[width=1\linewidth]{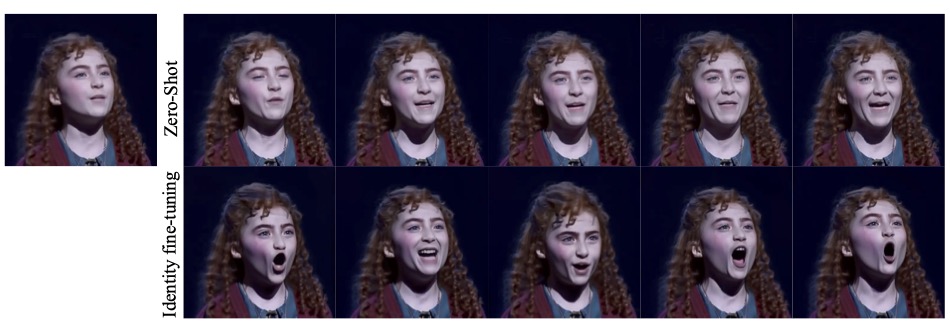}
    \caption{Visualization of identity fine-tuning personalization.
    Enhancing identity-specific expression and pose features through personalized data fine-tuning facilitates the generation of animations closely resembling the targeted identity.}
\label{fig:personal}
\end{figure}

\subsection{Qualitative Results}
\paragraph{Different Portrait Styles.}
The study depicted in Figures~\ref{fig:qualitative_portrait} delves into the impact of various portrait styles, including but not limited to sketching, painting, AI generated images, sculpture, and more, on video synthesis using the proposed methodology. 
The results underscore the versatility and robustness of the approach in generating a diverse range of visual and auditory outputs.

\paragraph{Different  Audio Styles.}
Figure~\ref{fig:qualitative_audio} presents qualitative outcomes concerning distinct audio styles. 
The findings demonstrate that our method can efficiently handle a range of audio inputs to generate high-fidelity and visually coherent videos that align seamlessly with the audio content.

\paragraph{Audio-Visual Cross Attention.}
Figure~\ref{fig:hadvs} illustrates the comparison of audio-visual cross attention between the original full and the proposed hierarchical audio-visual cross attention. 
It is observed that the proposed hierarchical audio-visual cross attention effectively achieves fine-grained and precise alignment between audio, lip motion, and facial expressions. 
This alignment enhances the relevance during training and enables more precise motion control based on audio in the inference process.

\paragraph{Head and Expression Motion Diversity.}
Figure~\ref{fig:diversity} presents a qualitative comparison of head and expression motion diversity between existing methodologies and our proposed approach. 
The analysis demonstrates that our method excels in generating animations characterized by enhanced diversity in expression and pose motion.

\paragraph{Identity Fine-tuning Personalization.}
Figure~\ref{fig:personal} presents the outcomes of fine-tuning diverse identities using personalized data specific to each identity. 
The analysis reveals that the proposed hierarchical audio-driven visual synthesis adeptly captures the distinctive features of individual identities, leading to personalized animations that closely reflect the unique traits of each identity following the fine-tuning procedure.

\begin{figure}[h]
    \centering
    \begin{minipage}{\textwidth}
        \centering
        \includegraphics[width=0.9\textwidth]{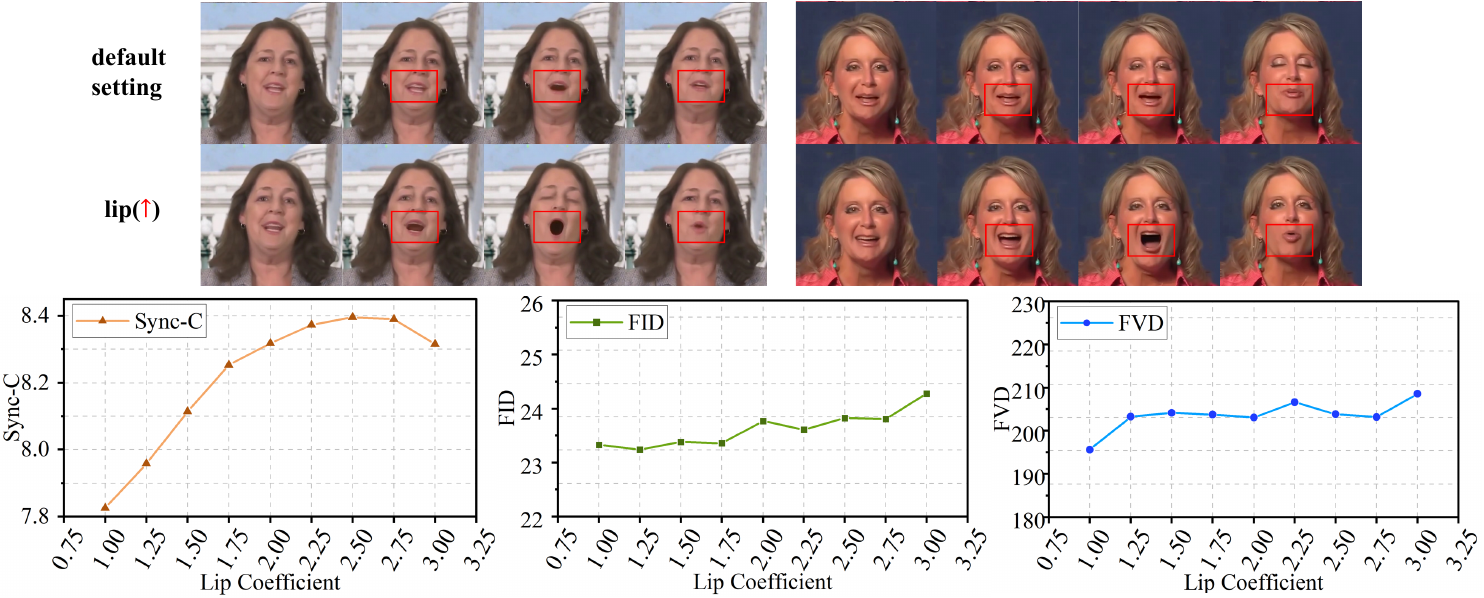}
        \vspace{-2mm}
        \label{fig:express_control_lip}
        \subcaption{Expression control by adjusting hierarchical weights.}    
        \vspace{2mm}
    \end{minipage}
    \begin{minipage}{\textwidth}
        \centering
        \includegraphics[width=0.9\linewidth]{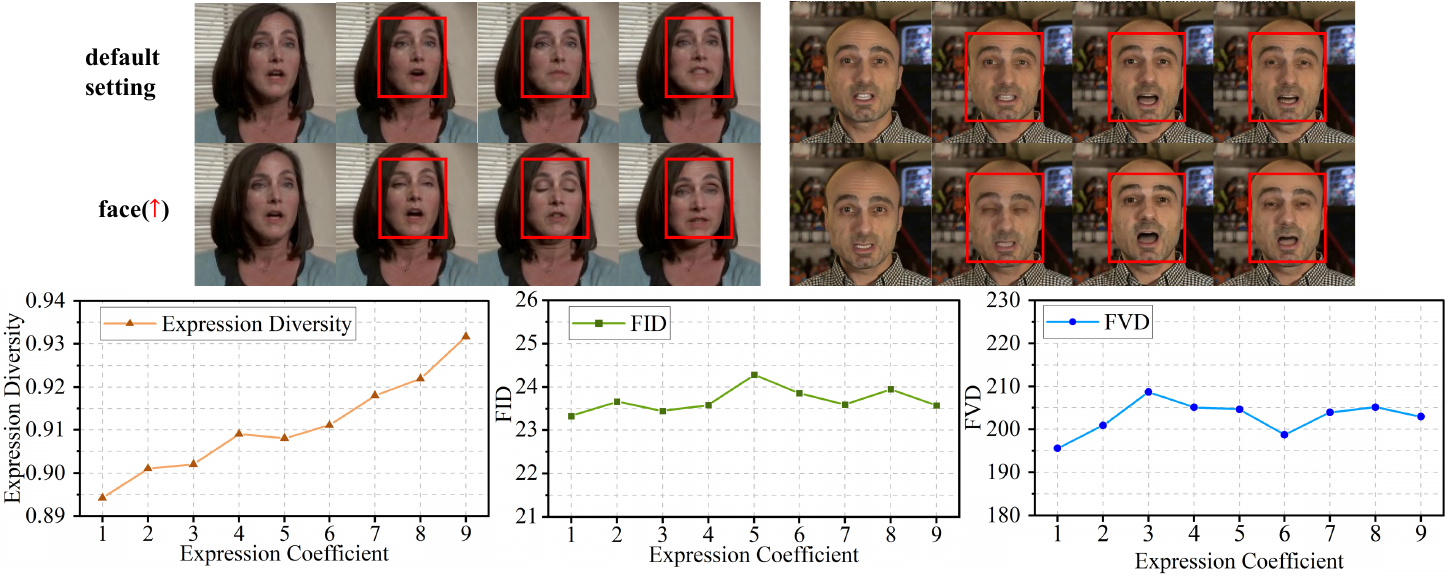}
        \vspace{-2mm}
        \label{fig:express_control_face}
        \subcaption{Qualitative comparison with applying higher coefficient on expression.}
        \vspace{2mm}
    \end{minipage}
    \begin{minipage}{\textwidth}
        \centering
        \includegraphics[width=0.9\linewidth]{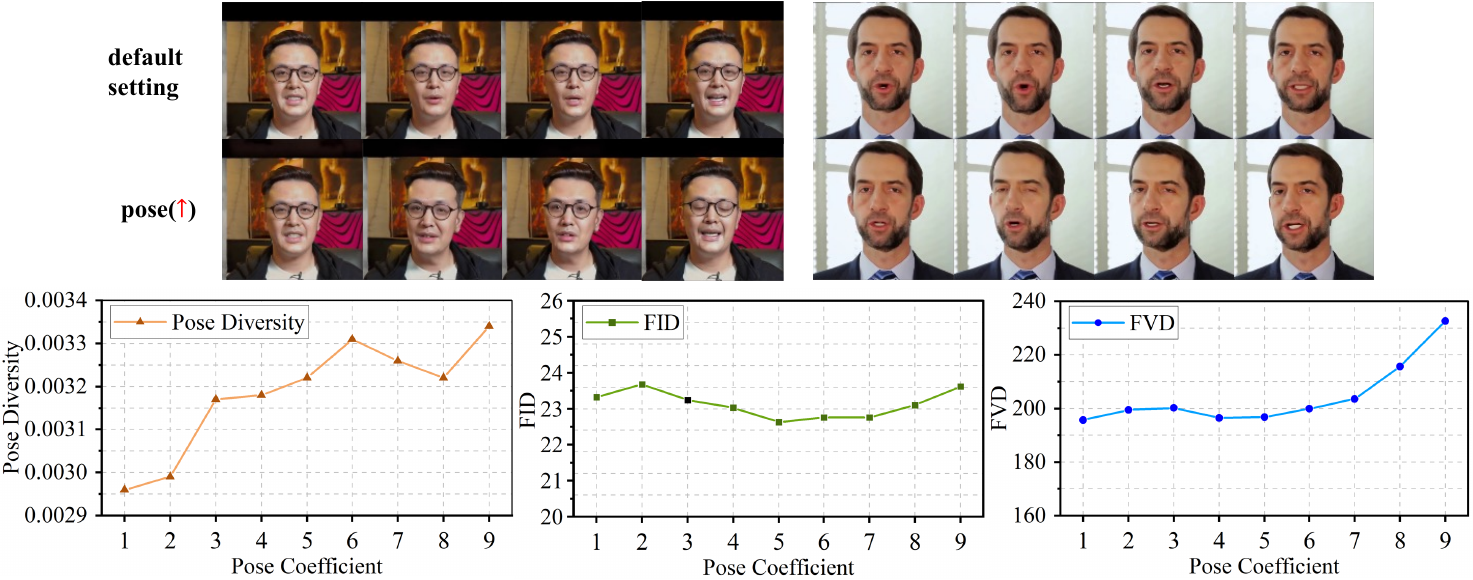}
        \label{fig:express_control_pose}
        \vspace{-2mm}
        \subcaption{Qualitative comparison with applying higher coefficient on pose.}
    \end{minipage}
    \caption{Motion control (pose, expression and lip) by adjusting hierarchical weights.
    he statistical analysis of these dynamics is conducted exclusively on the HDTF dataset to ensure consistency in the evaluation framework.}
    \label{fig:motion_control}
\end{figure}


\begin{table}[!t]
\centering
\begin{tabular}{cccc|ccccc}
\hline
Full & Lip           & Exp.         &  Pose & FID$\downarrow$ & FVD$\downarrow$ & Sync-C$\uparrow$ & Sync-D$\downarrow$ & E-FID$\downarrow$  \\ \hline
$\checkmark$   &              &              &             &  20.581    & 193.062   &  6.499   &  8.691    & 9.133  \\
$\checkmark$   & $\checkmark$ &              &             &  24.605    & 217.417   &  7.187   &  8.002    & 8.334   \\
$\checkmark$ &    & $\checkmark$ &             &  24.003     &  207.352  &  7.072   &  8.127    &  8.282  \\
$ \checkmark$   &   &  &$\checkmark$ &  23.452  &  205.636 &  6.436  &  8.502  &  8.375  \\
\hline
         & $ \checkmark$ & $\checkmark$ &$\checkmark$ &  \textbf{20.545}     &  \textbf{173.497} & \textbf{7.750}   &  \textbf{7.659}    & \textbf{7.951}  \\
\hline
\end{tabular}
\vspace{2mm}
\caption{Ablation study of hierarchical audio-visual (lip, expression, and pose) cross attention. 
The designation ``Full'' denotes the baseline configuration of full visual-audio cross attention. 
Subsequently, the study incrementally incorporates regional cross attention between audio and visual modalities, introducing lip, expression, and pose features individually. In our experimental setup, we adhere to the configuration specified in the last row of the table.
}
\label{tab:ablation_attention}
\end{table}

\begin{figure}[h]
    \centering
    \begin{minipage}{\textwidth}
        \centering
        \includegraphics[width=0.9\textwidth]{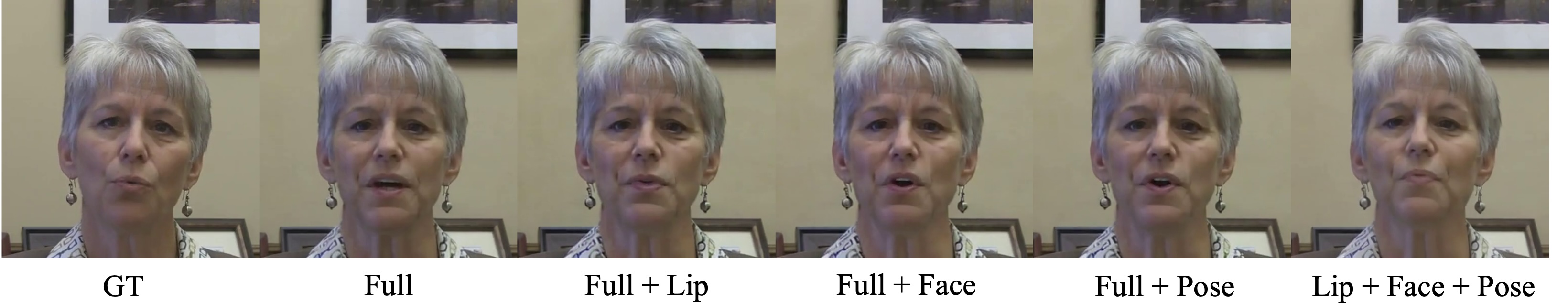}
        \vspace{-1mm}
        \label{fig:ablation_attention}
        \subcaption{Qualitative study of hierarchical audio-visual (lip, expression, and pose) cross attention.}    
        \vspace{2mm}
    \end{minipage}
    \begin{minipage}{\textwidth}
        \centering
        \includegraphics[width=0.9\linewidth]{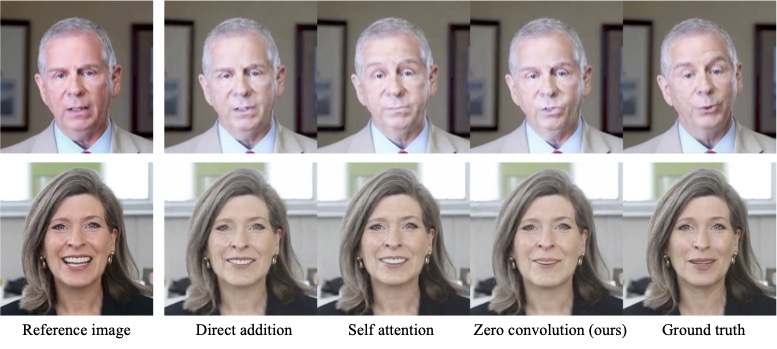}
        \vspace{-2mm}
        \label{fig:ablation_weight}
        \subcaption{Comparison of different attention weighting mechanisms in hierarchical audio-driven visual synthesis.}
        \end{minipage}
    \caption{Qualitative ablation study of (1) hierarchical audio-visual (lip, expression, and pose) cross attention; (2) different attention weighting mechanism in hierarchical audio-driven visual synthesis.}
    \label{fig:ablation}
\end{figure}

\subsection{Ablation Study}
\paragraph{Hierarchical Audio-Visual Cross Attention.}
The ablation study presented in Table~\ref{tab:ablation_attention} meticulously examines the impact of hierarchical audio-visual cross attention by systematically integrating lip, expression, and pose features. 
Analysis of the results reveals that the incorporation of lip features in isolation leads to enhancements in specific metrics, albeit with varying degrees of improvement across evaluation criteria. 
Furthermore, the combined inclusion of facial features with lip features yields mixed results but notably enhances synchronization confidence and reduces E-FID. Notably, the incorporation of all three modalities—lip, expression, and pose—exhibits the most substantial overall performance improvement, elevating the quality and coherence of audio-visual synthesis. 
This iterative refinement underscores the efficacy of hierarchical cross-attention across multiple features in achieving enhanced audio-visual integration.
Figure~\ref{fig:ablation}(a) complements this assessment with further qualitative insights into hierarchical audio-visual (lip, expression, and pose) cross attention.

\paragraph{Different Attention Weighting Mechanism.}
Table~\ref{tab:attention_mechanism} provides a quantitative assessment of various attention weighting mechanisms utilized in the hierarchical audio-driven visual synthesis module. 
Among the mechanisms examined, the ``Direct addition'' approach stands out with the lowest FID score of 19.580, indicating exceptional image quality. 
Remarkably, our ``Zero convolution'' strategy surpasses all others, demonstrating superior performance across critical metrics. 
Specifically, it achieves the highest FVD (173.497), Sync-C (7.750), Sync-D (7.659), and E-FID (7.951) scores, showcasing its robustness in generating synchronized, temporally coherent animations with utmost fidelity.
Figure~\ref{fig:ablation}(b) complements this analysis by providing additional qualitative comparisons of different attention weighting mechanisms in hierarchical audio-driven visual synthesis.

\begin{figure}[!t]
    \centering
     \includegraphics[width=0.9\linewidth]{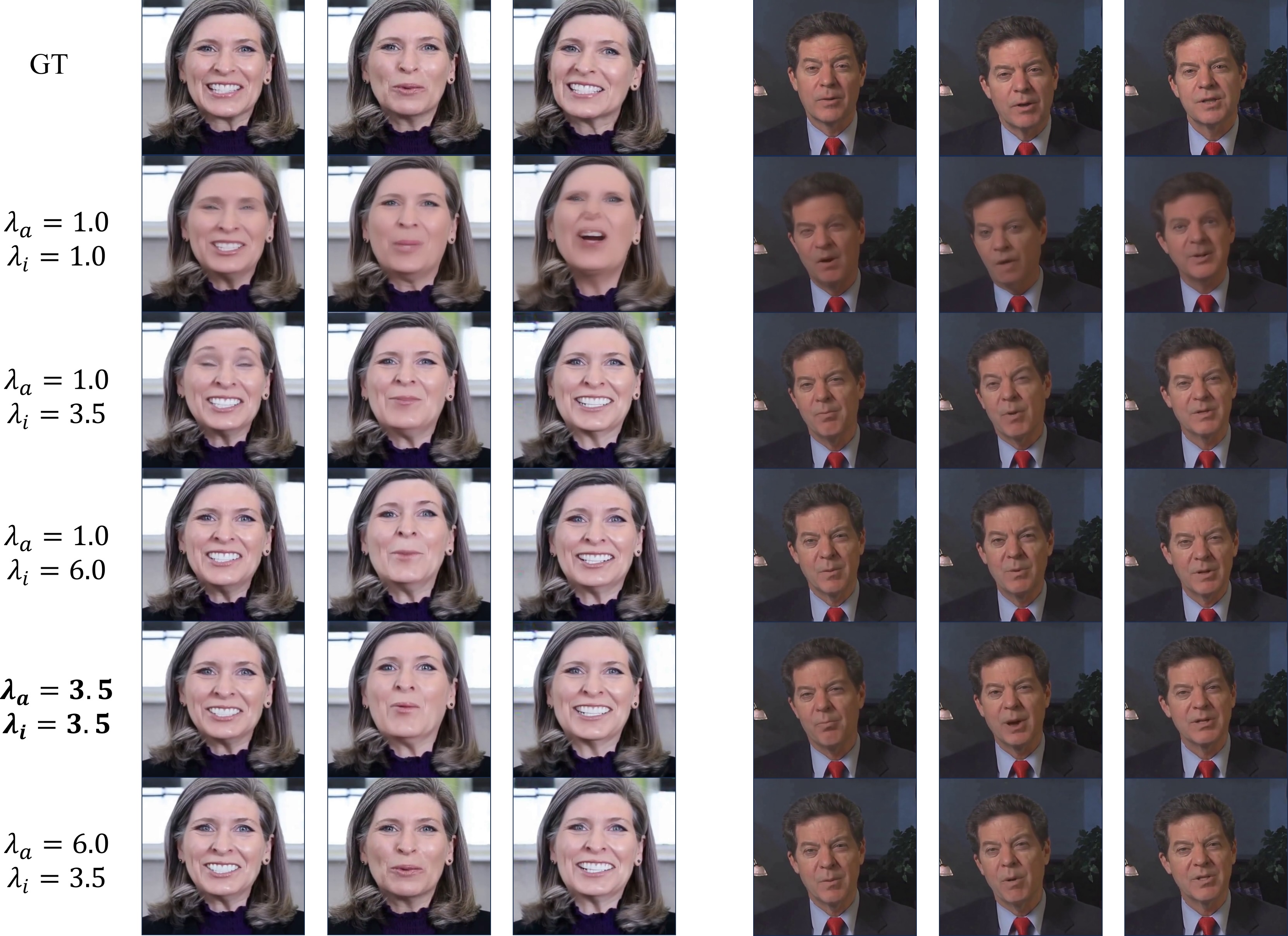}
    \caption{Qualitative study of audio and image CFG scales.}
    \label{fig:cfg}
\end{figure}

\begin{table}[!t]
\centering
\begin{tabular}{c|clclclclc}
\hline
Mechanism & FID$\downarrow$ & FVD$\downarrow$ & Sync-C$\uparrow$ & Sync-D$\downarrow$ & E-FID$\downarrow$  \\ \hline
Direct addition &  \textbf{19.580}    &  177.715   &   7.087  &  8.147   &   8.063   \\
Self attention   &  20.984   &  183.113   &   7.568   &  7.947    & 8.292       \\
Zero convolution (ours)  & 20.545 &   \textbf{173.497}  &   \textbf{7.750}  &    \textbf{7.659}  &     \textbf{7.951}   \\ \hline
\end{tabular}
\vspace{2mm}
\caption{Quantitative comparison of different attention weighting mechanism in the module of hierarchical audio-driven visual synthesis.}
\label{tab:attention_mechanism}
\end{table}

\paragraph{Expression Control by Adjusting Hierarchical Weights.}
Figure~\ref{fig:motion_control} illustrates the qualitative outcomes of expression control achieved through the manipulation of hierarchical weights.
The evaluation of various synthesis weight settings reveals a discernible trend: augmenting the lip weight leads to improved lip-sync accuracy while slightly compromising overall fidelity and video quality. 
In contrast, elevating the expression weight marginally enhances image and video quality but may have a slight adverse effect on synchronization metrics. 
Consequently, the selection of weight settings necessitates a judicious balance between fidelity, video quality, and synchronization, guided by the specific requirements of the application. 
For this study, we adhere to the default setting determined through the adaptive weighting process.

\begin{table}[!t]
\centering
\begin{tabular}{cc|cccccc}
\hline
audio & image & FID$\downarrow$ & FVD$\downarrow$ & Sync-C$\uparrow$ & Sync-D$\downarrow$ & E-FID$\downarrow$ \\ \hline
$\lambda_a$ = 1.0 &  $\lambda_i$ = 1.0 & 33.320 & 270.925 & 5.983  & 9.532 &    12.579    \\
$\lambda_a$ = 1.0 &  $\lambda_i$ = 3.5 & 23.210 & 194.295 & 5.742  & 9.435 &    8.565      \\
$\lambda_a$ = 1.0 &  $\lambda_i$ = 6.0 & 23.154 & 204.625 & 5.583  & 9.589 &     8.759   \\
$\lambda_a$ = 3.5 &  $\lambda_i$ = 3.5 & 23.167 & 195.179 & 7.658  & 7.894 & 7.951    \\
$\lambda_a$ = 6.0 &  $\lambda_i$ = 3.5 & 25.016 & 229.128 & 7.952  & 7.742 &    9.024      \\ \hline
\end{tabular}
\vspace{2mm}
\caption{Quantitative study of audio and image CFG scales. In our implementation, we adopt the setting ($\lambda_a = 3.5$ and $\lambda_i = 3.5$) for the balance of visual fidelity and motion diversity.}
\label{tab:cfg}
\end{table}

\paragraph{Audio and Image CFG Scales.}
Table~\ref{tab:cfg} presents a quantitative analysis of the generated videos under various configurations of audio and image CFG scales. 
Among the configurations, the setting of $\lambda_a = 3.5$ and $\lambda_i = 3.5$ achieves a balanced performance, with competitive FID (23.167), FVD (195.179), Sync-C (7.658), Sync-D (7.894), and E-FID (7.951) scores. 
This balance between visual fidelity and motion diversity underscores the effectiveness of this configuration in producing high-quality, synchronized videos.

\begin{table}[!t]
\centering
\begin{tabular}{c|ccccc}
\hline
Method          & GPU memory (GB) & Time (sec) \\ \hline
Inference w. HADVS  & 9.77 & 1.63  \\
Inference w.o. HADVS   & 9.76 & 1.63 \\
Inference ($256 \times 256$) & 6.62 & 0.46 \\
Inference ($1024 \times 1024$)  & 20.66 & 10.29 \\\hline
\end{tabular} 
\vspace{2mm}
\caption{Efficiency analysis of different steps and setting of the proposed approach. 
	      ``HADVS'' denotes the proposed hierarchical audio-driven visual synthesis.
	      ``$256 \times 256$'' and ``$1024 \times 1024$'' represent the inference video resoluton.} 
\label{tab:efficiency}
\end{table}

\paragraph{Efficiency Analysis.}
Table~\ref{tab:efficiency} presents an efficiency analysis of various configurations of our proposed approach. 
The inference with hierarchical audio-driven visual synthesis requires 9.77 GB of GPU memory and takes 1.63 seconds. In contrast, without hierarchical audio-driven visual synthesis, the GPU memory usage is slightly reduced to 9.76 GB while maintaining the same inference time of 1.63 seconds. 
Variations in video resolution significantly affect both GPU memory usage and inference time; processing at $256 \times 256$ resolution requires 6.62 GB of GPU memory and 0.46 seconds, whereas $1024 \times 1024$ resolution necessitates 20.66 GB of GPU memory and 10.29 seconds. 

\subsection{Limitation and Future Work}
Despite the advancements proposed in this study regarding portrait image animation, there exist several limitations that necessitate further exploration and consideration. 
These limitations underscore areas where future research endeavors can contribute to the refinement and augmentation of the presented methodology:
\textbf{(1) Enhanced visual-audio synchronization.} 
Future investigations could delve into more advanced synchronization techniques, potentially integrating sophisticated audio analysis methods or harnessing deeper cross-modal learning strategies. 
These advancements have the potential to refine the alignment of facial movements with audio inputs, particularly in contexts characterized by intricate speech patterns or nuanced emotional expressions.
\textbf{(2) Robust temporal coherence.} 
There is a need for further exploration of advanced temporal coherence mechanisms to address inconsistencies in sequences featuring rapid or intricate movements. 
Developing robust alignment strategies, possibly guided by long-term dependencies in sequential data or leveraging recurrent neural network models, could bolster frame-to-frame stability.
\textbf{(3) Computational efficiency.} 
It is imperative to optimize the computational efficiency of the diffusion-based generative model and UNet-based denoiser. 
Research into lightweight architectures, model pruning, or efficient parallelization techniques holds promise in rendering the approach more viable for real-time applications while minimizing resource utilization.
\textbf{(4) Improved diversity control.} 
The balance between the diversity of expressions and poses and the preservation of visual identity integrity remains a critical challenge. 
Future endeavors could concentrate on refining adaptive control mechanisms, potentially through the integration of more nuanced control parameters or sophisticated diversity metrics. 
Such refinements would ensure a more natural and diverse animation output while upholding the authenticity of facial identities.

\subsection{Social Risks and Mitigations}
In the context of the research presented in the paper, there are social risks associated with the development and implementation of portrait image animation technologies driven by audio inputs. 
One potential risk is the ethical implications of creating highly realistic and dynamic portraits that could potentially be misused for deceptive or malicious purposes, such as deepfakes. 
To mitigate this risk, it is essential to establish ethical guidelines and responsible use practices for the technology. 
Additionally, there may be concerns regarding privacy and consent related to the use of individuals' images and voices in the creation of animated portraits. 
Mitigating these concerns involves ensuring transparent data usage policies, obtaining informed consent, and safeguarding the privacy rights of individuals. 
By addressing these social risks and implementing appropriate mitigations, the research aims to promote the responsible and ethical development of portrait image animation technologies within the societal context.

\section{Conclusion}
This paper introduces a novel method for portrait image animation using end-to-end diffusion models, addressing challenges in audio-driven facial dynamics synchronization and high-quality animation generation with temporal consistency. 
The proposed hierarchical audio-driven visual synthesis module enhances audio-visual alignment through cross-attention mechanisms and adaptive weighting. 
By integrating diffusion-based generative modeling, UNet denoising, temporal alignment, and ReferenceNet, the method improves animation quality and realism.
Experimental evaluations demonstrate superior image and video quality, enhanced lip synchronization, and increased motion diversity, validated by superior FID and FVD metrics. 
The method allows flexible control over expression and pose diversity to accommodate diverse visual identities.

\bibliographystyle{ACM-Reference-Format}
\bibliography{nipsbib}

\end{document}